# A Multi-Plant Disease Diagnosis Method using Convolutional Neural Network

Muhammad Mohsin Kabir[1], Abu Quwsar Ohi[1], and M. F. Mridha[1]

[1]Bangladesh University of Business and Technology, Dhaka, Bangladesh
m97kabir2@gmail.com,quwsarohi@bubt.edu.bd,firoz@bubt.edu.bd

**Abstract.** A disease that limits a plant from its maximal capacity is defined as plant disease. From the perspective of agriculture, diagnosing plant disease is crucial, as diseases often limit plants' production capacity. However, manual approaches to recognize plant diseases are often temporal, challenging, and time-consuming. Therefore, computerized recognition of plant diseases is highly desired in the field of agricultural automation. Due to the recent improvement of computer vision, identifying diseases using leaf images of a particular plant has already been introduced. Nevertheless, the most introduced model can only diagnose diseases of a specific plant. Hence, in this chapter, we investigate an optimal plant disease identification model combining the diagnosis of multiple plants. Despite relying on multi-class classification, the model inherits a multi-label classification method to identify the plant and the type of disease in parallel. For the experiment and evaluation, we collected data from various online sources that included leaf images of six plants, including tomato, potato, rice, corn, grape, and apple. In our investigation, we implement numerous popular convolutional neural network (CNN) architectures. The experimental results validate that the Xception and DenseNet architectures perform better in multi-label plant disease classification tasks. Through architectural investigation, we imply that skip connections, spatial convolutions, and shorter hidden layer connectivity cause better results in plant disease classification.

## 1    Introduction

Plant disease is a common threat to the quality and quantity of global agricultural production. Disastrous plant disease enhances the current shortage of the food supply in which at least 0.8 billion people are inadequately fed [1]. Additionally, it is a significant threat to food security, where the number of consumers is increasing daily. To reduce harm, we must identify the disorder immediately. In particular, viral plant disorders have no solutions, and they spread rapidly. Thus, transited plants must be dispelled instantly to abstain from secondary infections. To remove infected plants immediately, the diagnosis of the infected plant is the most significant task.

The diagnosis and recognition of plant diseases play a vital role in ensuring the high quality and quantity of food production. Automated plant disease diagnosis is an inevitable research topic, and it is being studied with various researchers' approaches. The leaves of plants are the common element of plant disease detection, and the symptoms of most diseases start to appear on the leaves [2]. Therefore, identifying disease using leaf images is the most general method for researchers.

Farmers often use the internet or experienced gardeners' opinions to diagnose plant diseases. Sometimes, farmers take a small portion of an infected plant or picture it to the local agricultural health center. The diagnosis of plant disorders by manual optical observation of plant leaves is often time-consuming and laborious. Furthermore, inaccurate diagnosis is prevalent. In addition, experienced agriculturists and plant pathologists often fail to diagnose specific diseases and lead to mistaken solutions. They may sometimes be misdiagnosed because of the wide variety of disease symptoms, and these symptoms look very close to each other. Genetic testing is preferable, but it is expensive and time-consuming.

Therefore, an efficient and highly accurate disease diagnosis and recognition method for plant diseases must be introduced. Several research approaches have proposed automated plant disease diagnosis methods that include pattern recognition [3,4], machine learning [5], and deep learning [6]. Most of the techniques are plant-specific, and in some cases, they are disease-specific as well. Hence, multi-label classification is used in this work. Multi-label classifications can output several classes at the same time, from the same input image.

With the advancements of machine learning, computer vision applications have achieved enormous success. Success leads to implementing novel approaches and models, which now form a new class, known as deep learning (DL). DL techniques have been introduced in the agricultural domain, and they have gained massive popularity due to robustness. Researchers have invented convolutional neural networks (CNNs) that solve the pattern recognition problem associated with images. The transfer learning strategy has further enhanced the evolution of deep CNN-based architectures. In a transfer learning strategy, the CNN model is initially trained on a comparatively large dataset. The trained model is referred to as a pre-trained model. The pre-trained model further recognizes similar image patterns from the same or different domain of datasets. Transfer learning strategy often helps to avoid overfitting of DL architectures on small datasets. Furthermore, it also reduces the training iterations of DL architectures on the other datasets.

The overall contribution of the paper is summarized can be summarized as,

- We investigate a multi-label CNN classifier that can identify multiple plants and the related plant diseases. This is the first research endeavor that consolidates numerous plants' diagnosis to the best of our knowledge.
- We combine six different plants, including Potato, Tomato, Corn, Rice, Grape, and Apples for our experiment. The experimental dataset contained a total of 28 diseases of the six plants.
- We evaluate six popular image recognition baseline strategies that include Densenet, Inception, Mobilenet, ResNet, VGG, and Xception. Different implementations of these popular baselines are also included, resulting in a total of 15 baseline implementations. Further, we perform a rigorous investigation to point out the optimal architectural schemes.

The rest of this book chapter is organized as follows: Section 2 presents the related work. Section 3 introduces the dataset. In Section 4, the overall architecture of deep CNN is described. Section 5 contains the model's evaluation and compares the results of the architectures. Finally, Section 6 concludes the chapter.

## 2    Related work

With advancements in computer vision, progress has been achieved in the identification and diagnosis of plant diseases. Numerous diagnosis and identification techniques are proposed by the following image segmentation procedures, feature extraction, and pattern recognition. Before the evolution of deep learning, the popular classification approaches that were used for disease detection in plants include random forest [7], artificial neural network (ANN) [8], k-nearest neighbor (KNN) [9], and support vector machine (SVM) [5]. Recognition methods using the procedures mentioned before improved plant disorder diagnosis. However, these approaches depend on the extraction and selection of visible disease features. Recently, several works on automated plant disease diagnosis and identification have been developed using deep learning techniques.

Kawasaki et al. [6] proposed CNN architectures to recognize cucumber leaf disease and obtained 94.9% accuracy. CNN is the most useful classifier for image recognition in both small and large-scale datasets. It has shown excellent performance in image processing and classification [10]. Mohanty et al. [11] trained a deep learning model for recognizing 14 crop species and 26 crop diseases with 99.35% accuracy using GoogleNet and AlexNet architecture. CNN can perform both feature extraction and image classification. Srdjan et al. [12] proposed a plant disease recognition approach to classify healthy leaves and 13 different diseases based on CNNs. The results demonstrate that robust computing infrastructure makes CNN a suitable candidate for disease recognition.

However, some defects and difficulties include collecting a large labeled dataset, which is challenging. Although CNN gives much better accuracy due to large datasets' unavailability, transfer learning approaches have been introduced in plant disease classifications. Transfer learning consists of a pre-trained network where only the last classification levels' parameters need to be inferred to obtain the classification results [13].

To date, no multi-label approaches have been introduced to explore multiple plant disease identification. This work aims to construct a robust multi-label transfer learning approach that will identify the plant and its disease with very low computational complexity and higher accuracy.

In this chapter, we study transfer learning for deep CNNs for multi-label plant disease identification and diagnosis techniques. Transfer learning utilizes knowledge from source models to improve learning in the objective task. Transfer learning reduces both training iterations and data required to achieve better results. Furthermore, due to the knowledge transform often transfer learning strategies perform better generalization.

**Table 1.** This table reports the plants and types of infection that our collected dataset contains. The count of each type of infection is represented in the 'Samples' column.

| Plant Name | Condition | Samples |
|---|---|---|
| Tomato | Healthy | 1955 |
| | Early Blight | 1955 |
| | Late Blight | 1955 |
| | Leaf Mold | 1955 |
| | Septoria Leaf Spot | 1955 |
| | Spider Mites | 1955 |
| | Target Spot | 1955 |
| | Tomato Mosaic Virus | 1955 |
| | Yellow Leaf Curl Virus | 1955 |
| Potato | Healthy | 152 |
| | Early Blight | 152 |
| | Late Blight | 152 |
| Corn | Healthy | 2052 |
| | Cercospora Leaf Spot | 2052 |
| | Common Rust | 2052 |
| | Northern Leaf Blight | 2052 |
| Rice | Healthy | 1046 |
| | Brown Spot | 1046 |
| | Hispa | 1046 |
| | Leaf Blast | 1046 |
| Apple | Healthy | 2200 |
| | Apple Scab | 2200 |
| | Black Rot | 2200 |
| | Cedar Apple Rust | 2200 |
| Grape | Black Measles | 2115 |
| | Black Rot | 2115 |
| | Leaf Blight | 2115 |
| | Healthy | 2115 |

# 3 Dataset

Six publicly available plant disease datasets are used during the experiments. The datasets were gathered from Kaggle [14]. The dataset contains leaf images of 6 different plants where each plant has 4 to 9 related diseases. Table 1 shows how the database is distributed in terms of plant species and diseases. Additionally, the number of images used in each class is shown in Table 1. Fig. 1 shows the infected leaves of apple, namely, Apple scab, Black rot, and Cedar apple rust, with a healthy leaf image. Also, in Fig. 1, disorder images of corn are shown: Cercospora leaf spot, Common rust, and Northern leaf blight. This figure shows the Black measles, Black rot, and Leaf blight disease images of grape leaves. Early blight and Late blight are disordered potatoes, as shown in the figure. Three diseases of tomatoes, including Early blight, Late blight, Spider mites, are shown in Fig. 1.

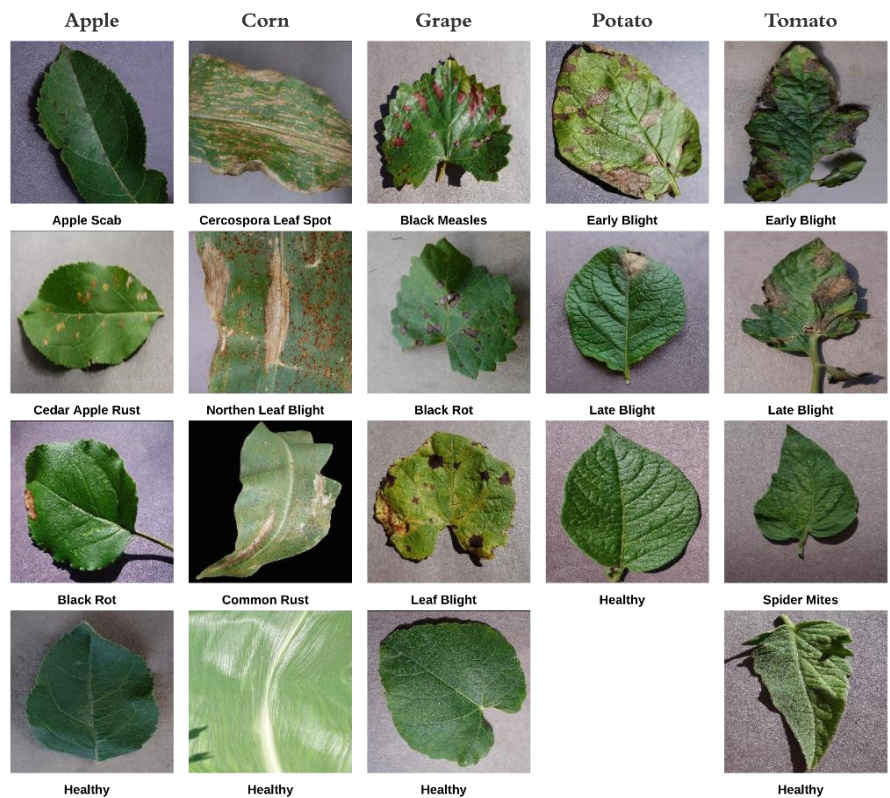

**Fig. 1.** This figure illustrates five plants and the types of diseases that each plant contains. Each row presents a set of leaf images caused by disease conditions for a specific plant.

## 4 Methodology

Different CNN architectures are implemented and benchmarked to perform multi-label classification. In the following sub-sections, the general process and layers of various CNN architectures are briefed.

### 4.1. Image pre-processing

As CNN architectures require input images to be of the same shape, each image data is reshaped into 120 by 120 pixels. Data normalization ensures that each input parameter has an analogous data distribution and results in faster convergence of the CNN. Therefore, each channel of the reshaped leaf images is normalized as,

$$Normalize(D) = \begin{bmatrix} d_{11} & \cdots & d_{1m} \\ \vdots & \ddots & \vdots \\ d_{n1} & \cdots & d_{nm} \end{bmatrix} / 255 \tag{1}$$

Where $D$ is the single-channel leaf image matrix, $n$ is the number of rows, and $m$ is the number of columns of the leaf image matrix.

### 4.2. Baseline architecture

The concept of computer vision is to enable machines to understand the world as humans do. CNN is a deep learning technique that takes input, adds weights and biases, and classifies images. Efforts have been made to explain the methodologies of CNN architectures [15].

In this book chapter, we focus on implementing the network architecture for robust plant disease diagnosis. Fig. 2 shows the CNN architecture, with the input layer (the raw image), convolutional layers, dense layer, and an output layer.

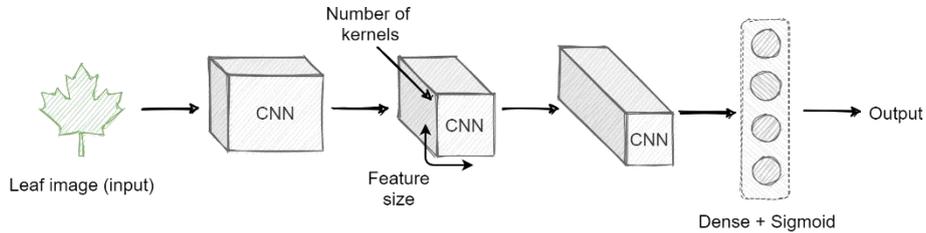

**Fig. 2.** The neural network architecture of plant diagnosis from leaf images. Each of the cubes represents an output of the convolution. The height and width are the gained information, and each cube's depth is equal to the number of kernels. Each convolution is followed by batch normalization and an activation layer. After the final convolution, it is converted into a linear set of nodes. Each node flows values through a sigmoid activation function.

**Input layer:** The inputs of the CNN architecture is the raw leaf images of different plants. Leaf images with different widths and heights are resized into the shape of $120 \times 120 \times 3$ before being given to the CNN architecture.

**Convolutional layers:** Multiple convolution layers follow the input layer of the model. It is a mathematical operation that takes two inputs, an image matrix and a filter (or kernel), and conserves the relationship between pixels by learning image features. In every layer, there is a bank of $m_1$ kernels. Each layer identifies particular features at every location on the input. The output $Y_i^l$ of layer $l$ consists of the $m_3^l$ feature of size $m_1^l \times m_2^l$. The $ith$ feature map denoted $Y_i^l$, is computed as,

$$Y_i^{(l)} = B_i^{(l)} + \sum_{j=1}^{m_1^{(l-1)}} K_{i,j}^{(l)} \times Y_j^{(l-1)} \tag{2}$$

Where $B_i^{(l)}$ is a bias matrix and $K_{i,j}^{(l)}$ is the filter of size. Using convolution reduces the computational parameter. A fully connected layer would contain at least the same number of trainable parameters the same as the number of pixels in an input image. Such a large number of parameters often result in overfitting. On the contrary, convolution does not increase the number of parameters. Instead, it only searches for a specific feature matrix $K$ that is learned through backpropagation. Recent deep neural network (DNN) architectures limit the height and width of the matrix $K$ to a maximum value of 5. Hence the trainable parameters are balanced, and therefore convolution is used in the first layer of the architecture.

**Batch normalization:** Batch normalization changes inter-layer outputs into a standard format. Batch normalization re-calibrates each of the data values based on the mean and variance for a specific data batch. Batch normalization increases the stability of DNN architectures and often leads to faster convergence. The general formula of normalization is defined as follows,

$$x'_i^{(k)} = \frac{x_i^{(k)} - \mu_B^{(k)}}{\sqrt{\sigma_B^{(k)^2}}} \tag{3}$$

Where $x'_i^{(k)}$ is the normalized value of the $kth$ hidden unit. $\mu_B^{(k)}$ is the mean value, and $\sigma_B^{(k)^2}$ is the variance of the $kth$ hidden unit. $B$ defines the data of a particular batch.

**ReLU:** ReLU function is used in every convolution for a simple calculation that returns the value provided as input directly. The function returns zero if it receives any non-positive input, but for any positive value of $x$, it returns that value. So it can be written as,

$$ReLU = \max(0, x) \tag{4}$$

The non-linearity of ReLU causes DL architectures to detect the optimal position properly.

**Dense layer:** The dense layer, also called a fully connected layer, works on a flattened input where every input is connected to all neurons. Assigning a dense layer is a simple way of sensing nonlinear combinations of the high-level features extracted using previous CNN layers. A single dense layer is used in the architecture to classify the disorder. Mathematically, a dense layer can be represented as,

$$d(x) = Activation(w^T x + b) \tag{5}$$

Here, $w = [w_1, w_2, \ldots, w_n]^T$ represents the weight vector of the dense layer, and $b$ represents the bias value of the dense layer. The activation function is a nonlinear function that defines the final output for a given input. A dense layer is used in the late part of the proposed deep neural network architecture to classify the plant species and type of the diseases. The sigmoid function is used as an activation function in dense layer.

**Sigmoid:** Sigmoid activation is used for multi-label classification of plant species identification and detecting disorder. It takes a real value as input and outputs another value [0,1]. In our architecture, it defines plant species identification and disorder diagnosis. A sigmoid function can be represented as follows,

$$\sigma(x) = \frac{1}{1+e^{-x}} \tag{6}$$

Here, $x$ represents the input of the sigmoid function. The sigmoid function is used as the final activation function represented in Eq. 6. Every node of the dense layer contains a sigmoid function for classifying the plant and disease.

### 4.3. Loss function

To calculate the loss, we have used binary cross-entropy in the architectures. The binary cross-entropy loss function calculates the loss of an input that is stated below,

$$L_{ce}(y, o) = -\sum_l(y_l \log o_l) - (1 - y_l) \times \log(1 - o_l) \tag{7}$$

Here, $y$ and $o$ are the target label and the output of the model, respectively. $y_l$ is the target for label $l$, and $o_l$ is the prediction for label $l$.

## 5    Evaluation

In this section, firstly, the evaluation metrics are defined. Later, the empirical setup is explained. Finally, we present the evaluation with a detailed analysis.

### 5.1. Evaluation metric

We have used accuracy, precision, and recall evaluation metrics based on the confusion matrix. A confusion matrix summarizes prediction results that measure the performance for machine learning, deep learning classification problems that contain four measures: true positive (TP), true negative (TN), false positive (FP), and false negative (FN). We evaluate the performance of the architecture by these measurements.

**Precision:** Precision defines all the positive classes the model predicted correctly; how many are actually positive. To obtain the value of precision, the total number of correctly classified positive examples are divided by the total number of predicted positive examples. The equation can be stated as,

$$Precision = \frac{TP}{TP+FP} \tag{8}$$

**Recall:** It defines how much the model predicted correctly among all positive classes. A recall is the ratio of the total number of correctly classified positive examples divided by the total number of positive examples. The equation can be stated as,

$$Recall = \frac{TP}{TP+FN} \tag{9}$$

**F1-score:** F1-score gives an overall estimation of the precision and recall of a test subject. It is the harmonic mean of the precision and recall of a test subject. Formally, F1-score can be defined as,

$$F1\_score = 2 \times \frac{Precision \times Recall}{Precision + Recall}$$

Each of the metrics results in a scale of [0, 1]. A higher score defines the better performance of a model.

### 5.2. Experimental setup

*Python* is used to collect data, pre-processing, experiments, and evaluations of the model [16]. The neural network architecture is implemented in *Keras* [17]. *Numpy* [18] is used to perform basic mathematical operations. Also, we have used *Keras* to implement deep learning models. *TensorFlow* [19] is used to generate the GPU execution of neural networks. For as a pre-training, we used *ImageNet* weights for each of the models. The input shape of the leaf images is $120 \times 120 \times 3$. The dataset has been split into train, test, and validation sets with a percentage of 50%, 25%, and 25%, respectively. The training dataset refers to the portion of data used to train the DL model in current literature. Whereas, the validation dataset is used to measure the perfection of the DL model while training. Furthermore, the testing data is considered the final set of data used to measure the model's performance on entirely unseen data. The Keras implemented *Adam* [20] optimization function is used to train the model for all the datasets with a learning rate of 0.001. A batch size of 128 is used for model training.

### 5.3. Experiments and comparisons

For proper evaluation, each model was pre-trained on the *ImageNet* dataset. Each of the results presented in this paper is presented as a mean of four runs. Each model is trained using the training dataset with a limit of 100 epochs. However, the training is halted if the loss on the validation dataset does not improve for ten epochs.

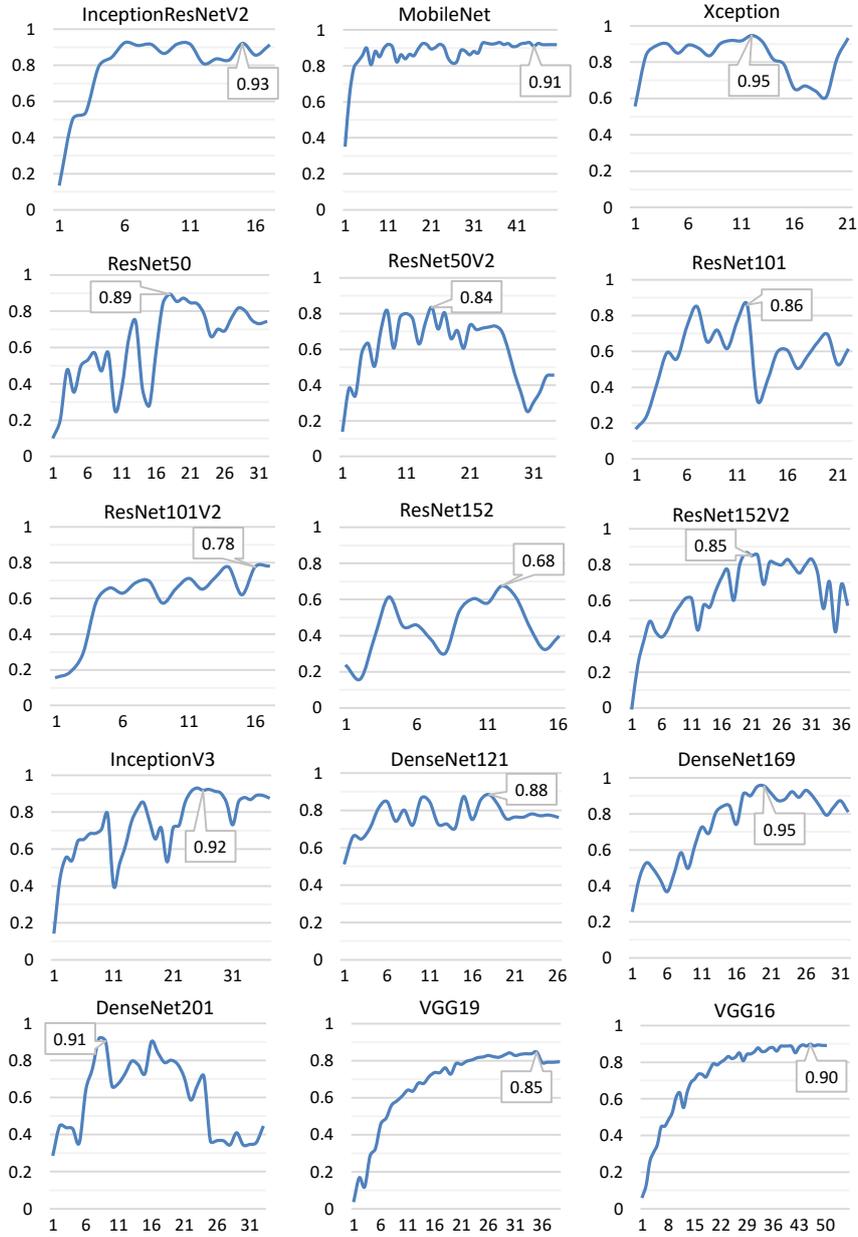

**Fig. 3.** For each CNN architecture, a separate graph is represented corresponding to the F1-score calculated on the validation dataset. Each graph's horizontal axis represents the epochs, and the vertical axis represents the F1-score in the scale of [0, 1]. The data illustrated are the mean of four runs.

Fig. 3. represents the F1-score on the validation dataset of the different models. From the presented graphs, it can be indicated that almost all of the architectures generate better F1-scores. However, due to overfitting in the training dataset., some specific ResNet architectures (ResNet50V2, ResNet101V2, and ResNet152) and VGG architectures hardly reach a minimum F1-score of 0.9. On the contrary, MobileNet architecture performs better compared to the trainable parameters available in the architecture. DenseNet, Xception, and Inception architectures avoid overfitting due to the skip connections. Among the various architectures, DenseNet169 and Xception reach the highest F1-score of 0.95.

**Table 2.** This table represents the precision, recall, and F1-score (in percentage) of different CNN architectures measured on the test dataset. Furthermore, the number of trainable parameters of each CNN implementation is also reported. The highest score is marked in bold.

| Model | Parameters (million) | Precision | Recall | F1-score |
|---|---|---|---|---|
| DenseNet121 [21] | 7 | 94.34 | 93.87 | 94.1 |
| DenseNet169 [21] | 13 | 97.87 | 96.85 | 97.36 |
| DenseNet201 [21] | 19 | 97.51 | 96.65 | 97.08 |
| InceptionV3 [22] | 23 | 95.9 | 95.19 | 95.55 |
| InceptionResNetV2 [23] | 54 | 96.92 | 93.95 | 95.41 |
| MobileNet [24] | 3 | 95.85 | 95.75 | 95.8 |
| ResNet50 [25] | 24 | 94.8 | 92.35 | 93.56 |
| ResNet50V2 [26] | 24 | 96.66 | 95.15 | 95.9 |
| ResNet101 [25] | 43 | 94.26 | 91.44 | 92.83 |
| ResNet101V2 [26] | 43 | 87.04 | 84.07 | 85.53 |
| ResNet152 [25] | 59 | 64.21 | 59.93 | 62.0 |
| ResNet152V2 [26] | 59 | 93.54 | 92.01 | 92.77 |
| VGG16 [27] | 137 | 92.53 | 89.62 | 91.05 |
| VGG19 [27] | 142 | 88.38 | 85.05 | 86.69 |
| Xception [28] | 21 | **97.88** | **96.9** | **97.38** |

The scores on the test data are reported in Table 2. The scores of each model are generated for the best weight found on the validation dataset. Xception performs better w.r.t. to the precision and recall on the test dataset. On the contrary, DenseNet169 slightly falls off from the maximum result. From the overall observation, it can be identified that Xception and DenseNet architectures perform better in the multi-label classification of the plant disease.

DensNet architectures contain massive parallel skip connections. Skip connections refer to the junction of some early layers through concatenation. Skip connections were first observed in ResNet architectures. Theoretically, it is implied that increasing the depth of a DL architecture should improve the classification performance. However, practically, it was not observed until the idea of skip connections was introduced.

DenseNet is designed based on the concept that convolution networks can perform better if they contain shorter linkage with the input layer and output layer. This method often helps architectures avoid overfitting and finding optimal combinations of neural activations for numerous input patterns.

On the contrary, Xception architecture is a combination of VGG and Inception architecture methodology. Furthermore, instead of general convolutions, Xception architectures perform spatial convolutions. Spatial convolutions work parallelly on multiple filters and tend to recognize texture features better. Xception also implements a skip connection strategy. However, it avoids heavy parallel connections, as observed in DenseNet architectures. Instead, Xception architecture implements at most a single pair of parallel flow due to the skip connection.

The overall benchmarks and evaluation illustrate the recent improvements in the DL architectures in computer vision. The result states that the recently investigated architectures mostly perform better in image recognition tasks. Also, the chapter validates the usefulness of skip connections, spatial convolutions of the DNN architectures.

## 6 Conclusion

This chapter implements and tests a multi-plant diagnosis method validated based on different image classifier baselines. We practiced a transfer learning scheme to train and test our approach precisely. Further, we evaluate the architecture on a dataset consisting of 6 different plants and 28 different diseases. We observe that spatial convolution, skip connections, and shorter hidden layer connectivity can massively improve the performance of multi-plant disease classification. As no dataset is available for multi-label disease classification, we will consider adding more diverse plants and diseases to evaluate our methods in future work. We strongly believe that this research work's contribution will be regarded as complementary to the present work, paving the way for significant research on transfer learning approaches for plant and disease identification.